# Improving the Interpretability of Support Vector Machines-based Fuzzy Rules

DUC-HIEN NGUYEN [a] and MANH THANH LE [b]
[a] *College of Information Technology, The University of DaNang*
[b] *Hue University*

**Abstract.** Support vector machines (SVMs) and fuzzy rule systems are functionally equivalent under some conditions. Therefore, the learning algorithms developed in the field of support vector machines can be used to adapt the parameters of fuzzy systems. Extracting fuzzy models from support vector machines has the inherent advantage that the model does not need to determine the number of rules in advance. However, after the support vector machine learning, the complexity is usually high, and interpretability is also impaired. This paper not only proposes a complete framework for extracting interpretable SVM-based fuzzy modeling, but also provides optimization issues of the models. Simulations examples are given to embody the idea of this paper.

## 1. Introduction

Fuzzy systems are known to a superior paradigm for processing the vague and uncertainty information and provide clear advantages in knowledge representation. The operations of fuzzy systems heavily depend on the inference process and the associated fuzzy rules in the inference engine. A number of methods to define fuzzy rule-based modeling have been discussed in the literature, including heuristic and data-driven methods [3, 5, 6, 7, 8, 10, 11, 12].

Extracting fuzzy models from Support Vector Machines (SVMs) is first investigated in [3]. In order to avoid local minimum problem SVMs utilize quadratic programming to find optimal separable plane, this concludes that the number of support vectors is proportional to the preciseness of the obtained model. In other words, while the performance of the derived model is acceptable, the interpretability is usually impaired.

In this paper, the relationship between SVMs and fuzzy systems is investigated. The relationships, especially the differences between these two models are emphasized. It is argued that the essential difference between SVMs and fuzzy systems is the interpretability, which enables fuzzy systems to be easily comprehensible. Based on the discussions on their relationships, a complete framework for extracting interpretable fuzzy rules from trained SVMs is proposed.

The rest of this paper is organized as follows. The relationships between SVMs and fuzzy rules and a proposed algorithm to establish interpretable fuzzy models from SVM are shown in Section 2. The effectiveness of this scheme is demonstrated through empirical evaluations in Section 3 followed by Conclusions in Section 4.

## 2. Extracting Interpretable Fuzzy Rules from Support Vector Machines

In this section, we first give an approach for extracting fuzzy rules from Support Vector Machine (SVM), and then explain how to optimize the fuzzy rules system and propose an algorithm for the conversion of SVM into interpretable fuzzy rules.

*2.1. Extracting fuzzy rules from support vector machine*

Support vector machine (SVM), which is proposed by Vapnik, is a new machine learning method based on the Statistical Learning Theory and is a useful technique for data classification [2]. SVM has been recently introduced as a technique for solving regression estimation problems [4], and has also been used in finding fuzzy rules from numerical data [3, 5, 10]. In the regression estimation task, the basic theory of SVM [2] can be briefly presented as follows:

Given a set of training data $\{(x_1, y_1), \dots, (x_l, y_l)\} \subset X \times \mathcal{R}$, where X denotes the space of input patterns. The goal of ε Support vector regression is to find a function $f(x)$ that has at most ε deviation from the actually obtained targets $y_i$ for all the training data, and at the same time is as flat as possible. That is, the errors would be ignored as long as they are less than ε, but any deviation larger than this would not be accepted.

For the case of nonlinear regression, the decision function has the form

$$f(x) = \sum_{i=1}^{l} (\alpha_i - \alpha_i^*) K(x_i, x) + b \tag{1}$$

Subject to

$$\sum_{i=1}^{l} (\alpha_i - \alpha_i^*) = 0, \text{ and } C \geq \alpha_i, \alpha_i^* \geq 0, \forall i, \tag{2}$$

Where, the constant C determines the trade-off of error margin between the flatness of $f(x)$ and the amount of deviation in excess of ε that is tolerated; and $K(x_i, x)$ is a kernel function defined as

$$K(x_i, x_j) = \langle \Phi(x_j), \Phi(x_j) \rangle \tag{3}$$

The input points $x_i$ with $(\alpha_i - \alpha_i^*) \neq 0$ are called support vectors. The difference to the linear case is that w is no longer given explicitly. Besides, the task becomes to find flattest function in feature space, not in the input space.

On the other hand, fuzzy rule-base that generally consists of set of IF-THEN rules is the core of the fuzzy inference [3, 6]. Suppose there are m fuzzy rules, it can be expressed as following forms:

$R_j$: IF $x_1$ is $A_1^j$ and $x_2$ is $A_2^j$ and ... and $x_n$ is $A_n^j$ THEN y is $B^j$, for $j = 1..M$ (4)

where $x_i (i = 1, 2, \dots n)$ are the input variables; y is the output variable of the fuzzy system; and $A_i^j$ and $B^j$ are linguistic terms characterized by fuzzy membership functions $\mu_{A_i^j}(x_i)$ and $\mu_{B^j}(y)$, respectively. If we choose product as the fuzzy conjunction operator, addition for fuzzy rule aggregation, and height defuzzification, then the overall fuzzy inference function is

$$f(x) = \frac{\sum_{j=1}^{M} \bar{z}^j \left( \prod_{i=1}^{n} \mu_{A_i^j}(x_i) \right)}{\sum_{j=1}^{M} \prod_{i=1}^{n} \mu_{A_i^j}(x_i)} \tag{5}$$

where $\bar{z}^j$ is the output value when the membership function $\mu_{B^j}(y)$ achieves its maximum value.

In order to let equation (1) and (5) are equivalent, at first we have to let the kernel functions in (1) and the membership functions in (5) are equal. The Gaussian membership functions can be chosen as the kernel functions since the Mercer condition [1] should be satisfied. Besides, the bias term b of the expression (1) should be 0.

While the Gaussian functions to be chosen as the kernel functions and membership functions, and the number of rules equals the number of support vectors, then (1) and (5) becomes

$$f(x) = \sum_{i=1}^{l}(\alpha_i - \alpha_i^*)\exp\left(-\frac{1}{2}\left(\frac{x_i - x}{\sigma_i}\right)^2\right) \tag{6}$$

and

$$f(x) = \frac{\sum_{j=1}^{l}\bar{z}^j\exp\left(-\frac{1}{2}\left(\frac{x_j - x}{\sigma_j}\right)^2\right)}{\sum_{j=1}^{l}\exp\left(-\frac{1}{2}\left(\frac{x_j - x}{\sigma_j}\right)^2\right)} \tag{7}$$

The inference of fuzzy systems can be modified as [3]

$$f(x) = \sum_{j=1}^{l}\bar{z}^j\exp\left(-\frac{1}{2}\left(\frac{x_j - x}{\sigma_j}\right)^2\right) \tag{8}$$

and the center of Gaussian membership functions are selected as

$$\bar{z}^j = (\alpha_i - \alpha_i^*) \tag{9}$$

Then, the output of fuzzy system (5) is equal to the output of SVM (1). However, the equivalence has some shortcomings: 1) the modified fuzzy model removes the normalization process; therefore, the modified fuzzy model sacrifices the generalization. 2) the interpretability cannot be provided during the modification.

An alternative approach is to set the kernel function of SVMs as

$$K(x_i, x) = \frac{\exp\left(-\frac{1}{2}\left(\frac{x_i - x}{\sigma_i}\right)^2\right)}{\sum_{i=1}^{l}\exp\left(-\frac{1}{2}\left(\frac{x_i - x}{\sigma_i}\right)^2\right)} \tag{10}$$

Consequently, the output of SVMs becomes

$$f(x) = \frac{\sum_{i=1}^{l}(\alpha_i - \alpha_i^*)\exp\left(-\frac{1}{2}\left(\frac{x_i - x}{\sigma_i}\right)^2\right)}{\sum_{i=1}^{l}\exp\left(-\frac{1}{2}\left(\frac{x_i - x}{\sigma_i}\right)^2\right)} \tag{11}$$

We only have to set the center of membership functions to $(\alpha_i - \alpha_i^*)$, then we can assure the output fuzzy systems (11) and the output of the SVMs (6) are equal. Notably, the expression (10) only can be achieved when the number of support vectors, l, is known previously.

For support vector machine for regression, when the number of SVs increases, we can get better regression line, but that, in turn make the interpretability and preciseness of the SVM-based fuzzy model becomes a trade-off. Next section we discuss the interpretability issues of fuzzy systems and present an algorithm to extract interpretable fuzzy modes from support vector machines.

## 2.2. The Proposed Algorithm for the Conversion of SVM into Understandable Fuzzy Rules

The idea of Interpretable is usually related to the ability of a model to express the behavior of the modeled system in an understandable way [11]. Interpretability is one of the most important features of fuzzy systems. In following, we define some important conditions for a fuzzy system to achieve its interpretability:

- **Completeness and Diversity:** The fuzzy partitions of all variables in the fuzzy system should be both complete and well distinguishable (i.e., distribution diversity). Besides, the number of fuzzy subsets in a fuzzy partition should also be limited. The completeness and distribution diversity condition makes it possible to assign a clear physical meaning to each fuzzy subset in a fuzzy partition. Usually, this leads to a small number of fuzzy subsets. A distance measure between neighboring fuzzy sets can be defined as

$$S(A_i, A_j) = \frac{\mathfrak{M}(A_i \cap A_j)}{\mathfrak{M}(A_i) + \mathfrak{M}(A_j) - \mathfrak{M}(A_i \cap A_j)} \quad (12)$$

where $\mathfrak{M}(A) = \int_{x \in X} A(x) dx$.

- **Expressive Efficiency:** The structure of the fuzzy system should be as compact as possible. That is, the number of linguistic variables associated with a fuzzy rule or a fuzzy constraint should be as small as possible. An efficient structure could be determined via input selection techniques [9].

- **Consistency.** Fuzzy rules in the rule base are consistent with each other and consistent with the available a priori knowledge. An inconsistency problem can be categorized as one of the following situations:
  o Two or more fuzzy predicates have almost the same facts but have different conclusions. For example, two rules defined on $A_1$ and $A_2$, respectively. The consequents of the two rules are $B_1$ and $B_2$. If $S(A_1, A_2) \gg S(B_1, B_2)$, then the two rules have almost the same facts but have different conclusions. This type of in consistency problem is frequently occurred in a data-driven fuzzy modeling approach.
  o Two or more fuzzy predicates have contradictory conclusions. For example, the consequent parts of fuzzy rules cannot happen simultaneously.

The central issue in converting an SVM into a fuzzy model is to ensure that the extracted fuzzy rules are interpretable. That is, the conditions described in the previous sections should be held. In order to convert an SVM into an interpretability fuzzy system, the following conditions should be satisfied:

- *The number of support vectors should be limited.* In Section 2.1 we have shown that the number of rules is identical to the number of support vectors. Therefore, the number of support vectors should be limited to assure an interpretable fuzzy model could be obtained.
- *Redundant support vectors should be removed.* If two or more SVs are located in a neighborhood, the interpreted fuzzy sets will have high similarity measure. However, high similarity measure may cause not only the diversity condition, but also the consistency condition is impaired.

The above conditions are necessary conditions on SVMs for extracting interpretable fuzzy systems. In the **figure 1**, an algorithm is provided such that during the extraction, the above conditions could be satisfied.

```
1: procedure MODELEXTRACTION(H, k, tol)
2:     ▷ H : data set
3:     ▷ k : similarity measure
4:     initialize: C, ε, σ, step
5:     while error > tol do
6:         f(x) = Σ_{i=1}^{l}(α_i − α_i*)K(x_i, x) + b ←—SVM(C, ε, σ)
7:         SV = {(α_i − α_i*) : (α_i − α_i*) ≠ 0, i ∈ {0, . . . , 1}}
8:         f(x) =InterpretabilityTest(SV, n, σ, k, H)
9:         Modify the Hessian (kernel) matrix as
10:        H' = [ D'  −D' ;  −D'  D' ],  D'_{ij} = ⟨φ(x_i),φ(x_j)⟩ / Σ_j⟨φ(x_i),φ(x_j)⟩
11:        error= E[ ||f(x) − H||^2 ]
12:        ε = ε+step
13:    end while
14:    σ_i(t + 1) = σ_i(t) + δε_{1,i} [ (x−c)^2/σ^3 exp(−(x−c)^2/(2σ^2)) ]
15:    c_i(t + 1) = c_i(t) + δε_{1,i} [ −(x−c)/σ^2 exp(−(x−c)^2/(2σ^2)) ].
16:    return f(x) = Σ_{i=1}^{l}(α_i−α_i*)K(x_i,x) / Σ_{i=1}^{l}(α_i−α_i*)
17: end procedure
18: procedure INTERPRETABILITYTEST(SV, n, σ, k, H)
19:    repeat
20:        Compute S^G(A_i, A_j) = σe^{−d^2/σ^2} / (2σ−σe^{−d^2/σ^2})
21:        Select A_i* and A_j* such that S^G(A_i*, A_j*) = max_{i,j}{S^G(A_i, A_j)}
22:        if S^G(A_i, A_j) > k then
23:            merge A_i, A_j to create a new fuzzy set A_k
24:            replace A_i = A_k and A_j = A_k.
25:        end if
26:    until no more fuzzy sets have similarity S^G(A_i, A_j) ≥ k
27: end procedure
```

**Figure 1.** The SVM-IF Algorithm.

The major idea underlying SVM-based fuzzy model learning is that during the SVM learning, the number of SVs and their positions can be determined as a basic structure of the fuzzy models. Besides, the kernel functions corresponding to each support vectors should be allowed to have different variances.

Inputs are data set H, similarity measure k, and the tolerance, tol, between obtained model and input data. Three parameters relating to the support vector learning should be specified at first, they are: C, ε, and σ. The procedure *ModelExtraction* can be separated into two parts. The first part includes Line 5 to Line 13, using a While loop to produce support vectors until the error measure exceeds the specified tolerance, *tol*. From Line 14 to 16 are the second part, which uses normalized kernel functions to generate normalized fuzzy system. The correctness of the procedure *ModelExtraction* is shown in what follows.

The obtained model has similarity measure $S^G(A_i, A_j) \leq k$ is obvious since in the procedure *InterpretabilityTest*, two fuzzy sets with similarity measure larger than k will be merged.

In order to obtain a set of optimal fuzzy with different variances, we can adopt such as gradient decent algorithms or GAs. We derive the following adaptive algorithm to update the parameters in the fuzzy membership functions:

$$\sigma_j(t+1) = \sigma_j(t) + \delta\varepsilon_{1,i}\left[\frac{(x-c)^2}{\sigma^3}\exp\left(-\frac{(x-c)^2}{2\sigma^2}\right)\right]$$

$$c_i(t+1) = c_i(t) + \delta\varepsilon_{1,i}\left[\frac{-(x-c)}{\sigma^2}\exp\left(-\frac{(x-c)^2}{2\sigma^2}\right)\right] \qquad (14)$$

Notably, during the SVM learning, parameters of kernel function should be specified previously. In other words, the variances of Gaussian functions are fixed. In order to achieve an optimal solution, we present an adaptive parameter specifying method and show how to estimate the initial values for iterative optimization.

## 3. Simulation examples

In this subsection, simulation studies on the modeling of the Mackey-Glass time series [7] are carried out to show the feasibility of the proposed method. The Mackey-Glass time series is described by

$$\dot{x} = \frac{ax(t-\tau)}{1+x^b(t-\tau)} - cx(t) \qquad (15)$$

where $\tau = 30$, $a = 0.2$, $b = 10$, and $c = 0.1$. One thound data samples are used in the simulation, 500 samples for training and the other 500 samples for test. The goal is to predict $x(t)$ using $x(t-1), x(t-2)$. That is to say, the system has two inputs and one output. The obtained model has 9 rules and the simulation result is shown in **Table 1** and **Figure2**.

**Table 1.** A comparison between different modeling approaches

$R_1$ :if $x(t-2)$ is Gaussmf(0.56, 0.48) and $x(t-1)$ is Gaussmf(0.52, 0.51) then $y$ is 1.12
$R_2$ :if $x(t-2)$ is Gaussmf(0.56, 0.48) and $x(t-1)$ is Gaussmf(0.66, 1.09) then $y$ is 1.08
$R_3$ :if $x(t-2)$ is Gaussmf(0.56, 0.48) and $x(t-1)$ is Gaussmf(0.53, 1.39) then $y$ is 0.97
$R_4$ :if $x(t-2)$ is Gaussmf(0.65, 1.07) and $x(t-1)$ is Gaussmf(0.52, 0.51) then $y$ is 1.32
$R_5$ :if $x(t-2)$ is Gaussmf(0.65, 1.07) and $x(t-1)$ is Gaussmf(0.66, 1.09) then $y$ is 0.94
$R_6$ :if $x(t-2)$ is Gaussmf(0.65, 1.07) and $x(t-1)$ is Gaussmf(0.53, 1.39) then $y$ is 1.11
$R_7$ :if $x(t-2)$ is Gaussmf(0.53, 1.37) and $x(t-1)$ is Gaussmf(0.52, 0.51) then $y$ is 1.11
$R_8$ :if $x(t-2)$ is Gaussmf(0.53, 1.37) and $x(t-1)$ is Gaussmf(0.66, 1.09) then $y$ is 1.09
$R_9$ :if $x(t-2)$ is Gaussmf(0.53, 1.37) and $x(t-1)$ is Gaussmf(0.53, 1.39) then $y$ is 0.98

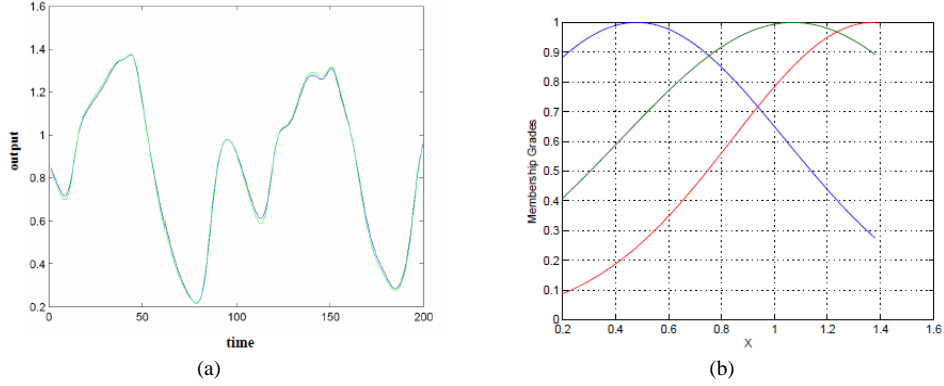

**Figure 2.** The final simulation result of the interpretable fuzzy model (RMSE = 0.0092)

Note that the centers and variances of these membership functions are adapted using the gradient decent methods. A comparison to traditional SVM, RBF NN, and ANFIS system can be found as shown in **Table 2**. The comparison shows that the proposed method is better than SVM, and similar to the results of RBF and zero order ANFIS.

**Table 2.** A comparison between different modeling approaches

| Number of Fuzzy Rules | Methods | | | | |
|---|---|---|---|---|---|
| | ANFIS-0 | ANFIS-1 | RBF | SVM | The Proposed SVM-FM |
| 170 | $< 10^{-10}$ | $< 10^{-10}$ | 0.0042 | 0.0540 | $< 10^{-10}$ |
| 6 | 0.0034 | 0.0023 | 0.0082 | 0.0509 | 0.0076 |
| 5 | 0.0041 | 0.0024 | 0.0086 | 0.0635 | 0.0090 |
| 4 | 0.0050 | 0.0030 | 0.0091 | 0.0748 | 0.0091 |
| 3 | 0.0074 | 0.0034 | 0.0141 | 0.1466 | 0.0092 |
| 2 | 0.0087 | 0.0041 | 0.0191 | 0.1955 | 0.0099 |

## 4. Conclusions

In this paper, we have presented an integrated approach to extracting interpretable fuzzy systems from support vector machines. The support vector learning mechanism provides a framework to extract support vectors for the use of fuzzy rules generation. The proposed approach resolved the problems of SVMs such as interpretability and complexity. The relationships between SVMs and interpretable fuzzy systems have also been discussed. Besides, the conditions for an interpretable fuzzy system were clarified. In order to extract interpretable fuzzy rules from an SVM, an adaptive ε-insensitive-parameter specifying algorithm has been introduced. We have shown that an SVM and a fuzzy system are not fully equivalent in terms of their semantic meanings and that the extraction of interpretable fuzzy rules from SVMs is both important and feasible for gaining a deeper insight into the local structure of the system to be approximated. Simulation examples were given to show the effectiveness of the proposed idea.